\documentclass{article}

\usepackage[final]{corl_2018}
\usepackage[dvipsnames]{xcolor}
\usepackage{graphicx}
\usepackage{tikz}
\usepackage{pgfplots}
\usepackage{subcaption}    
\usepackage{caption}
\usepackage{multirow}
\usepackage{footnote}
\usepackage[title]{appendix}
\usepackage[
  separate-uncertainty = true,
  multi-part-units = repeat
]{siunitx}
\pgfplotsset{compat=newest}
\usepgfplotslibrary{units}
\usepackage{amsmath,amssymb}
\usepackage{algorithm}
\usepackage[noend]{algpseudocode}
\usepackage{booktabs}
\usepackage{natbib}
\hypersetup{urlcolor={black}}
\usepackage[symbol]{footmisc}

\usepackage{caption} 
\pdfoutput=1
\title{RadGrad: Active learning with loss gradients}
\author{
    Paul Budnarain\begin{NoHyper}\thanks{Authors contributed equally.}\end{NoHyper}\\
    Department of Computer Science\\
    University of Toronto \\
    \texttt{paul.budnarain@mail.utoronto.ca} \\
    \And
    Renato Ferreira Pinto Junior\footnotemark[1]\\
    Department of Computer Science \\
    University of Toronto \\
    \texttt{renato@cs.toronto.edu} \\
    \And
    Ilan Kogan\footnotemark[1]\\
    Department of Statistical Sciences\\
    University of Toronto \\
    \texttt{mail@ilankogan.ca} \\
}

\makeatletter
\newenvironment{customlegend}[1][]{%
    \begingroup
    \pgfplots@init@cleared@structures
    \pgfplotsset{#1}%
}{%
    \pgfplots@createlegend
    \endgroup
}%
\def\addlegendimage{\pgfplots@addlegendimage}
\makeatother
\begin{document}
\maketitle
\begin{abstract}
Solving sequential decision prediction problems, including those in imitation learning settings, requires mitigating the problem of covariate shift. The standard approach, DAgger, relies on capturing expert behaviour in all states that the agent reaches. In real-world settings, querying an expert is costly. We propose a new active learning algorithm that selectively queries the expert, based on both a prediction of agent error and a proxy for agent risk, that maintains the performance of unrestrained expert querying systems while substantially reducing the number of expert queries made. We show that our approach, \texttt{RadGrad}, has the potential to improve upon existing safety-aware algorithms, and matches or exceeds the performance of DAgger and variants (i.e., SafeDAgger) in one simulated environment. However, we also find that a more complex environment poses challenges not only to our proposed method, but also to existing safety-aware algorithms, which do not match the performance of DAgger in our experiments.
\end{abstract}

\section{Introduction}
Sequential decision prediction problems, including imitation learning, differ from typical supervised learning tasks in that the actions of the agent affect the distribution of future observed states. The violation of the distributional stationarity assumption inherent in standard machine learning practice results in error compounding. As the agent drifts into states an expert would not have, error increases due to a lack of relevant training data. 

Consider, for instance, the task of teaching an autonomous car to stay within road boundaries. A facile approach would be to simply train a supervised learning system where environment states (e.g., road markings) are mapped to expert actions (e.g., the angle and velocity of a human driver) from a dataset of expert driving. During testing, if the car begins to drift off-course (inevitable for any algorithm that does not achieve perfect accuracy), the observed states would begin to differ from the training states. Compounding errors may cause the car to veer completely off-track.

To mitigate this issue, algorithms have been designed (i.e., DAgger \cite{ross2011reduction} and its derivatives) to iteratively aggregate training data on expert behaviour in states that the \textit{agent} visits. An underlying goal of these works is to maximize accuracy while minimizing the number of expert queries. Yet current approaches still require a high amount of expert input, making them infeasible for many real-world tasks. Consider teaching a robot surgeon: Having a surgeon demonstrate tens of thousands of surgeries is impractical. We propose a new algorithm that requires less expert input than DAgger while performing similarly. Our approach outperforms current state-of-the-art DAgger alternatives (i.e., SafeDAgger \cite{zhang2016query}) in query efficiency at a similar computational cost, increasing the breadth of real-world problems that can be solved with an imitation learning approach.

\section{Related Work}
We begin by introducing some notation to facilitate comparison of approaches and guide the rest of this paper. In the most general setting, we are given a set of expert demonstrations consisting of states $s$ and the corresponding expert actions (determined by the expert policy $\pi^{*}$): $\mathcal{D} = \{s_i, \pi^{*}(s_i)\}$. We seek to find a policy $\pi \in \Pi$ that closely mimics the expert policy $\pi^{*}$. We define a surrogate loss function that captures how ``close" the two policies are $\ell(\pi^{*}, \pi)$ and seek to minimize it:

\begin{equation}
\hat{\pi} = arg\,min_{\pi \in \Pi}\; \mathbb{E}_s [\ell(\pi^{*}, \pi)]  
\end{equation}

Unfortunately, we do not have knowledge of the underlying expert policy and instead have access only to its manifestation as a map from observed states to actions. Training only on states observed by the expert, as in supervised learning, is known to generally lead to poor performance due to covariate shift \cite{ross2011reduction}. We may improve our estimate of $\pi^{*}$, and thus potentially improve $\ell(\pi^{*}, \pi)$, by collecting additional expert demonstrations at cost $C(s_i)$ during agent-observed state $s_i$. Accordingly, we minimize subject to a maximum cost $\mathcal{C}$

\begin{equation}
min_{\pi \in \Pi}\; \mathbb{E}_s [\ell(\pi^{*}, \pi)] \; s.t. \; \sum_{i=1}^N C(s_i) < \mathcal{C}
\end{equation}

and must make a choice at each agent-observed $s_i$ whether we wish to query the expert. In the original DAgger algorithm \cite{ross2011reduction}, the cost $C(s_i)$ is implicitly assumed to be zero and the expert is always queried. A number of works have enhanced standard DAgger with probabilistic active learning machinery to determine when querying the expert is optimal under non-zero expert cost.

SafeDAgger \cite{zhang2016query} uses an initial set of demonstrations to train a binary risk classifier that predicts whether the agent will make a mistake in a given state, and then uses this classifier to choose when to query the expert. DropoutDAgger \cite{menda2017dropoutdagger} uses the Bayesian interpretation of neural networks with dropout to measure the epistemic uncertainty associated with a state. However, it uses this estimate only to guide action selection, while still querying the expert every time. BAgger \cite{cronrathbagger} incorporates these two ideas by directly modelling the agent's error with respect to the expert as a Gaussian Process or Bayesian Neural Network. Then, it obtains an empirical estimate of a percentile-based worst-case loss to decide whether to query the expert.

In pure reinforcement learning, Bayesian Q-learning \cite{dearden1998bayesian} and Bayesian Deep Q-learning \cite{azizzadenesheli2018efficient} learn a probabilistic $Q$ function that incorporates the agent's uncertainty about future rewards. However, the goal of these works is to achieve an optimal exploration-exploitation trade-off, and they do not address how agents could benefit from access to expert demonstrations.

\begin{figure}[h]
    \centering
    \includegraphics[height=2in]{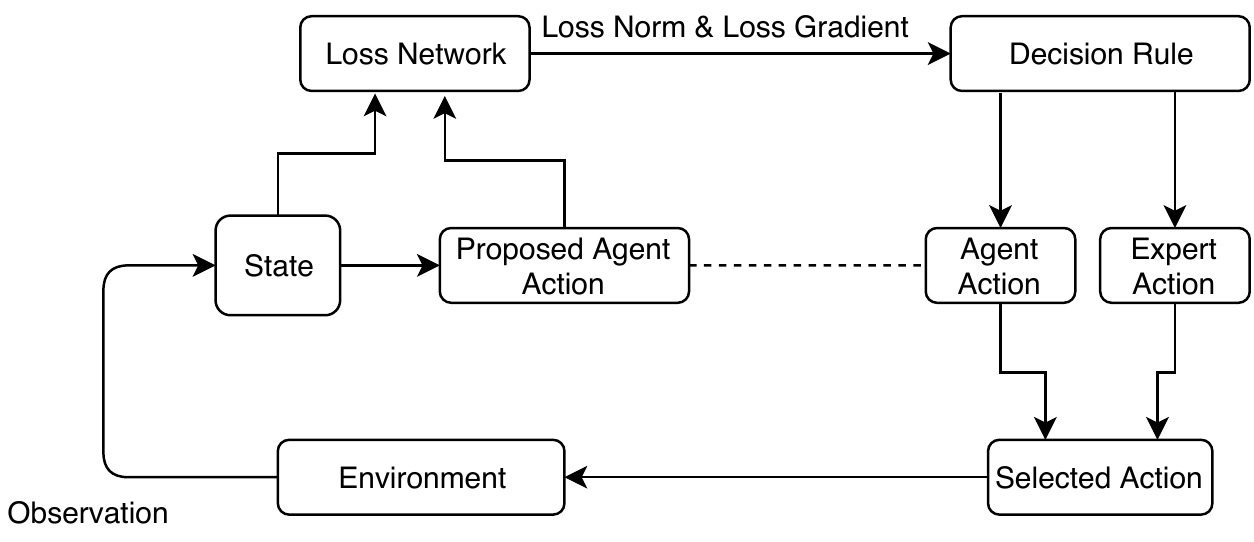}
    \caption{The \texttt{RadGrad} algorithm queries the expert both when the proposed agent action is predicted to be far from the expert's action, or if the gradient of this error is high with respect to the state and proposed action.}
    \label{fig:flowchart}
\end{figure}

Unlike other approaches, \texttt{RadGrad} introduces the concept of a loss gradient (Figure \ref{fig:flowchart}). SafeDAgger estimates whether a proposed agent action will exceed the unknown expert action beyond a safety threshold $\tau$ using what we term a loss network. We assume that the loss network is differentiable and query the expert both when the threshold is exceeded but also when the norm of the gradient of the prediction with respect to the concatenated vector of state and proposed action exceeds a separate threshold $\epsilon$. This gradient is a crude proxy for risk. As we describe later, this differs from the concept of uncertainty that Bayesian approaches and ensemble methods are well-suited for.

\section{Method}
\subsection{RadGrad Algorithm}
The four parts of our \texttt{RadGrad} approach (the primary network, the loss network, the loss gradient, and data aggregation) are summarized in Algorithm \ref{alg:radgrad} and detailed below.

\paragraph{Primary Network}
To find a policy $\pi \in \Pi$ that minimizes $\ell(\pi^{*}, \pi)$, we require a way to express $\pi$. Accordingly, we learn a function that maps from states in $\mathbb{R}^m$ to actions in $\mathbb{R}^n$. We employ a feed-forward neural network with hidden layer sizes $[128, 128, 32, 8]$ and dropout rate of $0.2$, trained using our set of expert demonstrations $\mathcal{D}$. At test time, an observed state $s_i$ is inputted into this function to generate a proposed agent action $\hat{a}_i$.

\paragraph{Loss Network}
We additionally learn a differentiable function, which we term a \textit{loss network}, that maps state and proposed agent action pairs in $\mathbb{R}^{m+n}$ into an estimate of the difference between the proposed agent action and the unknown expert action in $\mathbb{R}^n$. We employ a feed-forward neural network with hidden layer sizes $[128, 128, 64, 64, 32, 32, 16, 16, 8]$ and dropout rate of $0.2$. When the norm of the loss network output exceeds a safety threshold $\tau$, the expert is queried for expert action $a^{*}_i$. This is the strategy specified in Algorithm \ref{alg:radgrad}. An alternative implementation of the loss network, more similar to SafeDAgger, maps state and proposed agent action pairs in $\mathbb{R}^{m+n}$ to the probability that the norm of the loss network output exceeds the safety threshold $\tau$. In the latter case, the expert is queried for $a^{*}_i$ if the predicted probability exceeds $\frac{1}{2}$.

\paragraph{Loss Gradient}
Additionally, we calculate the norm of the gradient of the output of the loss network with respect to its input. If the norm exceeds a threshold $\epsilon$, then we query the expert for $a^{*}_i$. This norm is a proxy for risk. A large norm implies that a small change in either the state or proposed action would have a large impact on the probability of exceeding the threshold $\tau$, and thus the agent is in as, we define, a \textit{risky state} with high potential for error. Without the computationally-costly endeavour of building an ensemble or Bayesian neural network to measure uncertainty proper, we have built a proxy for measuring risk (which we treat as distinct from uncertainty).

\paragraph{Data Aggregation}
Whenever the expert is queried, we append the state and expert action pair $(s_i, a^{*}_i)$ to $\mathcal{D}$. In this work, we choose to take the expert action whenever we query the expert, although more fine-grained rules could be explored in future work. Appending these state-action pairs serves to shift the distribution of training states from those an expert would see to those the agent sees. We retrain the primary network on this new, aggregated dataset to improve agent performance, separating $20\%$ of the dataset for validation so as to reduce the risk of overfitting.

\begin{algorithm}
  \caption{\texttt{RadGrad} (Loss Gradient algorithm)}
  \begin{algorithmic}[1]
    \Procedure{RadGrad}{}
      \State Initialize $\mathcal{D} \gets \emptyset$
      \State Initialize $\pi_{agent, \: 1}$
      \State Initialize loss network $l_{agent, \:1}$
      \For{iteration $k = 1 : M$}
      \For{epoch $j = 1 : N$}
      \State Initialize environment and agent
      \For{timestep $i = 1 : T$}
      \State Observe state $s_i$
      \State $\hat{a}_i \gets \pi_{agent,\:k}(s_i)$
      \State $\hat l \gets l_{agent,\:k}(s_i, \hat{a}_i)$
      \If{$\hat l > \tau$ or $||\frac{\partial \hat{l}}{\partial [s_i; \hat{a}_i]}|| > \epsilon$}
        \State Query the expert to obtain $a^*_i \gets \pi^*(s_i)$
        \State Execute $a^*_i$
        \State $\mathcal{D} \gets \mathcal{D} \cup \{(s_i, \hat{a}_i, \hat{a}^*_i)\}$
      \Else
        \State Execute $\hat{a}_i$
      \EndIf
      \EndFor
      \EndFor
      \State Train $\pi_{agent,\:k+1}$ and $l_{agent,\:k+1}$ on $\mathcal{D}$
      \EndFor
      \State \textbf{return} best $\pi_{agent,\:k}$ on validation set
    \EndProcedure
  \end{algorithmic}
  \label{alg:radgrad}
\end{algorithm}

\subsection{Experimental Setup}
\paragraph{Algorithms}
We compare the performance of five primary algorithms in our analysis. These five include three non-gradient algorithms (DAgger, SafeDAgger, and Loss Network) and two gradient algorithms (SafeDAgger Gradient and Loss Network Gradient, or \texttt{RadGrad}). The non-gradient algorithms query the expert and execute the returned action if the loss network threshold is surpassed; the gradient algorithms query the expert in this case as well, but also if the gradient threshold is exceeded. SafeDAgger and SafeDAgger Gradient refer to a loss network that outputs the probability of surpassing $\tau$, as opposed to an output in $\mathbb{R}^n$.

Additionally, we consider the performance of three baseline methods: expert actions, supervised learning, and random selection. The expert action baseline is the reward achieved by the expert on the task. We consider the expert baseline only implicitly; we present loss measures as the difference in reward between the agent algorithm and the expert. We present the results of a simple supervised learning algorithm (which trains on expert demonstrations in expert-observed states only) to display the issue of covariate shift we wish to resolve. Finally, we present the random selection baseline. Random selection queries and follows the expert at random agent-observed states. A random selection baseline is necessary to establish the complexity of active learning approaches is warranted.

Finally, because we observe that random selection performs quite well in practice and hypothesize that an unbiased sampling strategy can be beneficial for convergence and stability, we test two hybrid algorithms: Loss Gradient Random (\texttt{RadGrad} Random) and SafeDAgger Gradient Random. At each timestep a fair coin is flipped to determine whether to use the loss-based versus random strategy.

\paragraph{Environment}
We test our approach in the Reacher-v2 and Hopper-v2 OpenAI gym environments \cite{brockman2016openai}. In Reacher-v2, a robotic arm with two degrees of freedom rotates to reach a randomly-positioned target. This environment maps from $s_i \in \mathbb{R}^{11}$ to $a_i \in \mathbb{R}^2$. In Hopper-v2, a two-dimensional one-legged robot hops as quickly as possible towards a target. The Hopper-v2 environment maps from $s_i \in \mathbb{R}^{11}$ to $a_i \in \mathbb{R}^{3}$. We selected open source environments for easy reproducibility.

\paragraph{Hyperparameters}

Table \ref{tab:hyper} summarizes the hyperparameters we used in our evaluations. These were chosen so as to optimize expert query efficiency while maintaining convergence to the algorithm's best policy. The random baseline hyperparameter was chosen so as to make that strategy competitive with active learning strategies in query efficiency (Equation \ref{eq:effic}).

\begin{table}[h]
    \centering
    \begin{tabular}{ccccc}
        \toprule
        Algorithm & Hyperparameter & Reacher-v2 & Hopper-v2 \\
        \midrule
        Loss & $\tau$ & 0.02 & 0.3\\
        Loss Gradient (\texttt{RadGrad}) & $\epsilon$ & 0.002 & 0.2  \\ \midrule
        SafeDAgger & $\tau$ & 0.04 & 0.3  \\
        SafeDAgger Gradient & $\epsilon$ & 1 & 200  \\ \midrule
        Random & $P(\text{Query})$ & $30\%$ & $30\%$ \\ \bottomrule
    \end{tabular}
    \caption{Hyperparameters for proposed algorithms and baselines. $\tau$ is the threshold on the norm of predicted loss, and $\epsilon$ is the threshold on the norm of the gradient of predicted loss with respect to input and action space. Note the two values of $\tau$ for Reacher-v2 differ since displayed values were individually-optimal.}
    \label{tab:hyper}
\end{table}

\section{Results}
Our results show that gradient-based methods can outperform their non-gradient-based counterparts in that they may yield higher rewards with only a modest increase in the number of expert queries required. To compare algorithm performance, we define the \textit{query efficiency} of estimated policy $\hat{\pi}$:

\begin{equation}
\text{Efficiency}(\hat{\pi}) \propto \frac{\text{reward}_{\hat{\pi}} - \text{reward}_{supervised}}{\sum_{i \in \mathcal{D}_{\hat{\pi}}} C(s_i)}
\label{eq:effic}
\end{equation}

This is the difference in loss between a supervised learning policy (trained only on expert actions in expert-observed states) and the active learning policy in question, divided by the total cost of querying the expert at states $s_i$ during the estimation of $\hat{\pi}$. For our purposes, we let $C(s_i) = 1 \: \forall \: s_i$, and thus $\sum_{i \in \mathcal{D}_{\hat{\pi}}} C(s_i)$ is simply the number of times the expert was queried in the estimation of $\hat{\pi}$ (i.e. $\#\mathcal{D}_{\hat{\pi}}$). Accurate policies that require few queries to estimate are query efficient.

Table \ref{tab:performance} shows the test-time performance, that is, average reward when expert demonstrations are not available, of all algorithms in each environment. It also shows, for each setting, the number of expert queries used to estimate the policy and the resulting efficiency.

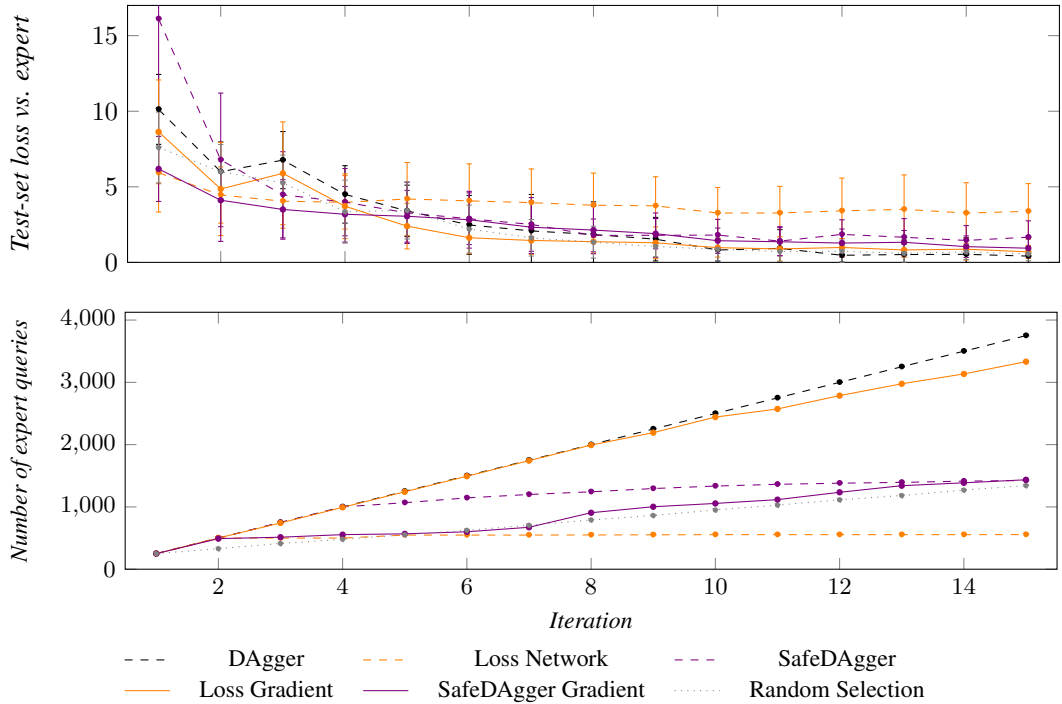
\begin{figure}[h!]
\centering
\begin{subfigure}{\textwidth}
    \centering
    \begin{tikzpicture}
    \begin{axis}[
        width = \textwidth,
        height = 5cm,
        unbounded coords=discard,
        ylabel={\textit{Test-set loss vs. expert}},
        xmin = 0.5, xmax = 15.5, ymin = 0.0, ymax=17.0,
        xticklabels={,,}
        font={\footnotesize},
        yticklabel style={
        text width=2.5em,align=right
        },
        yticklabel={
                    \pgfmathprintnumber[
                        fixed,
                        precision=0,
                        zerofill,
                    ]{\tick}
              }]
    
    \addplot[mark=*, dashed, mark size = 1, error bars/.cd, y dir=both, y explicit, error bar style = solid]
        table [x=iteration, y=loss, y error=err, col sep=comma]{reacher_final_dagger_for_plot.csv};
    \addplot[mark=*, dashed, mark size = 1, color = orange, error bars/.cd, y dir=both,y explicit, error bar style = solid]
        table [x=iteration, y=loss, y error=err, col sep=comma]{reacher_final_loss_002_for_plot.csv};
    \addplot[mark=*, dashed, mark size = 1, color = violet, error bars/.cd, y dir=both, y explicit, error bar style = solid]
        table [x=iteration, y=loss, y error=err, col sep=comma]{reacher_final_safedagger_004_for_plot.csv};
     
     \addplot[mark=*, solid, mark size = 1, color = orange, error bars/.cd, y dir=both, y explicit , error bar style = solid]
        table [x=iteration, y=loss, y error=err, col sep=comma]{reacher_final_gradient_loss_002_0002_for_plot.csv};
     \addplot[mark=*, solid, mark size = 1, color = violet, error bars/.cd, y dir=both, y explicit , error bar style = solid]
        table [x=iteration, y=loss, y error=err, col sep=comma]{reacher_final_safedagger_gradient_004_1_for_plot.csv};

   \addplot[mark=*, dotted, mark size = 1, color = gray, error bars/.cd, y dir=both, y explicit , error bar style = solid]
        table [x=iteration, y=loss, y error=err, col sep=comma]{reacher_final_random_30_for_plot.csv};
        \end{axis}
    \end{tikzpicture}
\end{subfigure}%

\begin{subfigure}{\textwidth}
    \centering
    \begin{tikzpicture}
    \begin{axis}[
        width = \textwidth,
        height = 5cm,
        unbounded coords=discard,
        xlabel={\textit{Iteration}},
        ylabel={\textit{Number of expert queries}},
        xmin = 0.5, xmax = 15.5, ymin = 0,
        font={\footnotesize},
        yticklabel style={
        text width=2.5em,align=right
        },
        scaled y ticks=false]
        
    \addplot[mark=*, dashed, mark size = 1]
        table [x=iteration, y=total_obs, col sep=comma]{reacher_final_dagger_for_plot.csv};
    \addplot[mark=*, dashed, mark size = 1, color = orange]
        table [x=iteration, y=total_obs, col sep=comma]{reacher_final_loss_002_for_plot.csv};
    \addplot[mark=*, dashed, mark size = 1, color = violet]
        table [x=iteration, y=total_obs, col sep=comma]{reacher_final_safedagger_004_for_plot.csv};
     
     \addplot[mark=*, solid, mark size = 1, color = orange ]
        table [x=iteration, y=total_obs, col sep=comma]{reacher_final_gradient_loss_002_0002_for_plot.csv};
     \addplot[mark=*, solid, mark size = 1, color = violet]
        table [x=iteration, y=total_obs, col sep=comma]{reacher_final_safedagger_gradient_004_1_for_plot.csv};

   \addplot[mark=*, dotted, mark size = 1, color = gray ]
        table [x=iteration, y=total_obs, col sep=comma]{reacher_final_random_30_for_plot.csv};
    \end{axis}
    \end{tikzpicture}
\end{subfigure}
\begin{tikzpicture}
    \begin{customlegend}[legend columns=3,legend style={draw=none,column sep=2ex, font=\small},legend entries={DAgger, Loss Network, SafeDAgger, Loss Gradient, SafeDAgger Gradient, Random Selection}]
    \addlegendimage{black,dashed,fill=black!50!red,sharp plot}
    \addlegendimage{orange,dashed,fill=black!50!red,sharp plot}
    \addlegendimage{violet,dashed,fill=black!50!red,sharp plot}
    \addlegendimage{orange,solid,fill=black!50!red,sharp plot}
    \addlegendimage{violet,solid,fill=black!50!red,sharp plot}
    \addlegendimage{gray,dotted,fill=black!50!red,sharp plot}
    \end{customlegend}
\end{tikzpicture}

\caption{A comparison of active learning approaches to DAgger in the \textit{Reacher-v2} task. DAgger yields rewards most similar to the expert, but gradient-based approaches perform competitively while reducing expert queries. Error bars are based on 100 trials per iteration and indicate $\pm$ one standard deviation.}
\label{fig:reacher}
\end{figure}

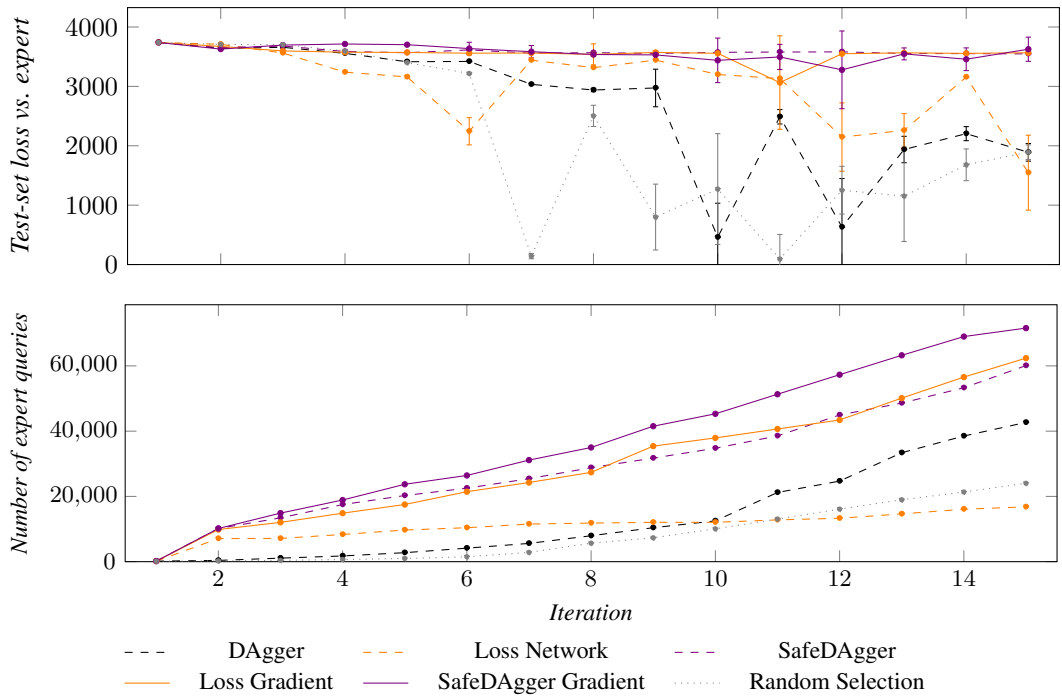
\begin{figure}[h!]
\centering
\begin{subfigure}{\textwidth}
    \centering
    \begin{tikzpicture}
    \begin{axis}[
        width = \textwidth,
        height = 5cm,
        unbounded coords=discard,
        ylabel={\textit{Test-set loss vs. expert}},
        xmin = 0.5, xmax = 15.5, ymin = 0.0,
        xticklabels={,,}
        font={\footnotesize},
        yticklabel style={
        text width=2.5em,align=right
        },
        yticklabel={
                \ifdim\tick pt=0pt
                    \pgfmathprintnumber[
                        fixed,
                        precision=0,
                        zerofill,
                    ]{\tick}
                \else
                    \num[
                        round-mode=figures,
                        round-precision=2,
                    ]{\tick}
                \fi}]
    
    \addplot[mark=*, dashed, mark size = 1, error bars/.cd, y dir=both, y explicit, error bar style = solid]
        table [x=iteration, y=loss, y error=err, col sep=comma]{hopper_final_dagger_for_plot.csv};
    \addplot[mark=*, dashed, mark size = 1, color = orange, error bars/.cd, y dir=both,y explicit, error bar style = solid]
        table [x=iteration, y=loss, y error=err, col sep=comma]{hopper_final_loss_03_for_plot.csv};
    \addplot[mark=*, dashed, mark size = 1, color = violet, error bars/.cd, y dir=both, y explicit, error bar style = solid]
        table [x=iteration, y=loss, y error=err, col sep=comma]{hopper_final_safedagger_03_for_plot.csv};
     
     \addplot[mark=*, solid, mark size = 1, color = orange, error bars/.cd, y dir=both, y explicit , error bar style = solid]
        table [x=iteration, y=loss, y error=err, col sep=comma]{hopper_final_gradient_loss_03_02_for_plot.csv};
     \addplot[mark=*, solid, mark size = 1, color = violet, error bars/.cd, y dir=both, y explicit , error bar style = solid]
        table [x=iteration, y=loss, y error=err, col sep=comma]{hopper_final_safedagger_gradient_03_200_for_plot.csv};

   \addplot[mark=*, dotted, mark size = 1, color = gray, error bars/.cd, y dir=both, y explicit , error bar style = solid]
        table [x=iteration, y=loss, y error=err, col sep=comma]{hopper_final_random_30_for_plot.csv};
        \end{axis}
    \end{tikzpicture}
\end{subfigure}%

\begin{subfigure}{\textwidth}
    \centering
    \begin{tikzpicture}
    \begin{axis}[
        width = \textwidth,
        height = 5cm,
        unbounded coords=discard,
        xlabel={\textit{Iteration}},
        ylabel={\textit{Number of expert queries}},
        xmin = 0.5, xmax = 15.5, ymin = 0,
        font={\footnotesize},
        yticklabel style={
        text width=2.5em,align=right
        },
        scaled y ticks= false]
        
    \addplot[mark=*, dashed, mark size = 1]
        table [x=iteration, y=total_obs, col sep=comma]{hopper_final_dagger_for_plot.csv};
    \addplot[mark=*, dashed, mark size = 1, color = orange]
        table [x=iteration, y=total_obs, col sep=comma]{hopper_final_loss_03_for_plot.csv};
    \addplot[mark=*, dashed, mark size = 1, color = violet]
        table [x=iteration, y=total_obs, col sep=comma]{hopper_final_safedagger_03_for_plot.csv};
     
     \addplot[mark=*, solid, mark size = 1, color = orange ]
        table [x=iteration, y=total_obs, col sep=comma]{hopper_final_gradient_loss_03_02_for_plot.csv};
     \addplot[mark=*, solid, mark size = 1, color = violet]
        table [x=iteration, y=total_obs, col sep=comma]{hopper_final_safedagger_gradient_03_200_for_plot.csv};

   \addplot[mark=*, dotted, mark size = 1, color = gray ]
        table [x=iteration, y=total_obs, col sep=comma]{hopper_final_random_30_for_plot.csv};
    \end{axis}
    \end{tikzpicture}
\end{subfigure}
\begin{tikzpicture}
    \begin{customlegend}[legend columns=3,legend style={draw=none,column sep=2ex, font=\small},legend entries={DAgger, Loss Network, SafeDAgger, Loss Gradient, SafeDAgger Gradient, Random Selection}]
    \addlegendimage{black,dashed,fill=black!50!red,sharp plot}
    \addlegendimage{orange,dashed,fill=black!50!red,sharp plot}
    \addlegendimage{violet,dashed,fill=black!50!red,sharp plot}
    \addlegendimage{orange,solid,fill=black!50!red,sharp plot}
    \addlegendimage{violet,solid,fill=black!50!red,sharp plot}
    \addlegendimage{gray,dotted,fill=black!50!red,sharp plot}
    \end{customlegend}
\end{tikzpicture}

\caption{A comparison of active learning approaches to DAgger in the \textit{Hopper-v2} task. DAgger and random sampling most reliably converge to competitive performance. Error bars are based on 100 trials per iteration and indicate $\pm$ one standard deviation.}
\label{fig:hopper}
\end{figure}

Figures \ref{fig:reacher} and \ref{fig:hopper} show test-time performance and training-time number of expert queries used over the course of training iterations for the two environments, Reacher-v2 and Hopper-v2, respectively. At each point in training, the current model is deployed on a batch of random test-time environments to generate the curves of performance over time shown in the graphs.

\begin{table}[h]
\centering
\begin{tabular}{@{}lllcllc@{}}
\toprule                                                                           
                           & \multicolumn{3}{c}{\textbf{Reacher-v2}}                                               &    \multicolumn{3}{c}{\textbf{Hopper-v2}} \\
\midrule                                                                           
                   \textbf{Algorithm} &  \textit{Queries} &           \textit{Loss} &  \textit{Efficiency}  &  \textit{Queries} &           \textit{Loss} &  \textit{Efficiency}\\
\midrule                                                                           
                  SafeDAgger (SD) &   1424 &  $1.67 \pm 1.08$    &             -6.3  &   60094 &    $3547 \pm 5$ &              22 \\
        SD Gradient &   1436 &  $0.94 \pm 0.62$                  &             -1.2  &   71591 &  $3626 \pm 203$ &              8 \\
        SD Gr. Random &   1551 &  $0.57 \pm 0.46$           &              1.3  &   53542 &   $606 \pm 843$ &             574 \\
\midrule
                       Loss &    556 &  $3.38 \pm 1.83$          &             -47  &   16786 &  $1547 \pm 631$ &            1270 \\
                       Loss Gradient &   3332 &  $0.70 \pm 0.43$ &              0.2  &   62378 &    $3567 \pm 4$ &              18 \\
                       Loss Gr. Random &   2200 &  $0.50 \pm 0.30$ &       1.2  &   64053 &  $2342 \pm 605$ &             209 \\
\midrule
                     DAgger &   3750 &  $0.41 \pm 0.62$ &              0.9  &   42682 &  $1890 \pm 148$ &             419 \\
                     Random &   1343 &  $0.56 \pm 0.46$ &              1.5  &   24025 &  $1892 \pm 122$ &             744 \\
                 Supervised &   3750 &  $0.77 \pm 0.57$ &              0  &  140164 &   $3679 \pm 19$ &              0 \\                                                                                                        
\bottomrule
\end{tabular}
\caption{Comparison of performance and query efficiency of gradient-based and non-gradient approaches. Displayed loss is loss in increment of expert loss, along with intervals of one standard deviation over 100 trials. Expert policies are obtained from Berkeley's Deep Reinforcement Learning course materials (\url{https://github.com/berkeleydeeprlcourse/homework/tree/master/hw1/experts}). Results presented are from final iteration of fifteen, with 100 trials at each iteration.}
\label{tab:performance}
\end{table}

We make three conclusions. First, adding gradient logic to safety-aware baselines (SafeDAgger Gradient and Loss Gradient) improves performance and efficiency on Reacher-v2, and integrating gradient logic with random sampling (SafeDAgger Gradient Random and Loss Gradient Random) further improves average reward as well as efficiency. This result suggests the validity of loss gradients as a proxy for risk, as well as the benefit of unbiased expert sampling. We further note that, although not statistically significant, test-time performance of risk-aware algorithms appears superior to that of DAgger early on. We hypothesize that this occurs because, early in training, risk-aware strategies shift the distribution of training data toward riskier states, causing the trained models to give more importance to those states than DAgger would.

Second, although DAgger shows best performance overall, random sampling is a strong baseline, converging to similar performance as DAgger with substantially improved query efficiency in both Reacher-v2 and Hopper-v2. This indicates that an unbiased sampling strategy may be a competitive model against which proposed active learning strategies should be tested.

Third, most safety- and risk-aware algorithms fail to converge to DAgger performance in the more complex environment Hopper-2. In this setting, DAgger and random sampling stand out as strong algorithms despite their simplicity. While some proposed algorithms (Loss and SafeDAgger Gradient Random) show promising performance and efficiency numbers, the fact that other safety-aware algorithms, including the established SafeDAgger baseline, fail to converge to DAgger performance makes us cautious to make strong conclusions from these data. While it is possible that these algorithms would converge to DAgger performance under more extensive hyperparameter tuning, this result hints at the challenges posed by richer environments to algorithms that aim at outperforming DAgger and random baselines.

A final observation is that in our Reacher-v2 simulations, SafeDAgger Gradient queries the expert fewer times than SafeDAgger proper for much of the training course (Figure \ref{fig:reacher}), even though the condition for querying the expert in SafeDAgger Gradient is more relaxed. We hypothesize that this occurs because SafeDAgger Gradient, by using the gradient of loss as a proxy for risk and obtaining expert demonstrations in the face of such risk, is better able to reduce future risk and thus future need for expert queries. We note, however, that this is not the case for Hopper-v2.

\section{Limitations}
While our work suggests the value of gradients as a proxy for risk in active learning, our experiments are hardly conclusive. Most notably, we did not complete an extensive analysis of the value of gradients in all of the major DAgger derivatives and instead focused our efforts on SafeDAgger.

Due to computational limitations, even though we deployed each trained agent at each training iteration in multiple randomly sampled test-time environments, we executed this procedure only once per algorithm. In other words, only one agent was trained per algorithm. For a more robust evaluation, we would train a number of agents for each algorithm to produce uncertainty estimates for the number of expert queries made as well.

While our Reacher-v2 simulations show that adding gradient logic to active learning decision rules has the potential to improve performance and, thus, should be further investigated, we could not replicate those results in the second, more challenging environment Hopper-v2. Not only did our gradient-based methods not improve performance over DAgger and random selection in Hopper-v2, but even the established SafeDAgger did not converge to a competitive policy in that case. While we do not rule out that more extensive hyperparameter search could improve the performance of those algorithms, we believe this result should be a call for more robust methods that can more easily be transferred to new environments.

Similarly, we believe further investigation of the trade-offs of active learning and unbiased random strategies to be necessary. Not only did we find that a random querying strategy is highly competitive to both DAgger and safety-aware strategies, but most importantly, we also found that random selection was the most robust policy when replicating our experiments on a new environment. Thus, unbiased strategies seem to provide stronger practical guarantees of generalization compared to current state-of-the-art active learning strategies. Of course, purely random sampling may not be possible due to safety risk; in these cases observation weighting may offer a compromise.

\section{Conclusion}
DAgger is able to improve over a purely supervised learning approach by mitigating the problem of covariate shift, but does so at a high expert querying cost. Various DAgger derivatives have been created to limit the number of expert queries made while maintaining similar policy quality as DAgger. We have shown that these methods may be possible to improve by incorporating gradients.

We experienced difficulty in replicating the performance of both popular active learning strategies and our proposed methods in a more complex environment. Further research on the robustness of active learning algorithms across environments is necessary.

Finally, we observed that a random selection algorithm, which obtains unbiased samples of expert demonstrations, is a strongly competitive alternative to both query-intensive and safety-aware methods. Future imitation learning active learning algorithms should compare to a random querying baseline to establish algorithmic complexity is warranted.

\newpage

\bibliography{output.bbl}
\end{document}